\documentclass[conference]{IEEEtran}
\IEEEoverridecommandlockouts
\usepackage{cite}
\usepackage{amsmath,amssymb,amsfonts}
\usepackage{algorithmic}
\usepackage{graphicx}
\usepackage{textcomp}
\usepackage{xcolor}
\usepackage{subcaption}
\usepackage{multirow}

\def\BibTeX{{\rm B\kern-.05em{\sc i\kern-.025em b}\kern-.08em
    T\kern-.1667em\lower.7ex\hbox{E}\kern-.125emX}}

\newcommand{\ie}{\emph{i.e.}\@{\!\@}{}}
\newcommand{\eg}{\emph{e.g.}\@{\!\@}{}}

\begin{document}

\title{Multi-object Video Generation 
from \\ Single Frame Layouts}
\makeatletter
\newcommand{\linebreakand}{%
  \end{@IEEEauthorhalign}
  \hfill\mbox{}\par
  \mbox{}\hfill\begin{@IEEEauthorhalign}
}
\makeatother
\author{\IEEEauthorblockN{Yang Wu\IEEEauthorrefmark{1}, Zhibin Liu\IEEEauthorrefmark{2}, Hefeng Wu\IEEEauthorrefmark{3} and Liang Lin\IEEEauthorrefmark{4}}
\IEEEauthorblockA{\IEEEauthorrefmark{1}\IEEEauthorrefmark{2}\IEEEauthorrefmark{3}\IEEEauthorrefmark{4}School of Computer Science and Engineering, 
Sun Yat-sen University\\
\IEEEauthorrefmark{3}\IEEEauthorrefmark{4}GuangDong Province Key Laboratory of Information Security Technology, 
Sun Yat-sen University\\
\{\IEEEauthorrefmark{1}wuyang36, \IEEEauthorrefmark{2}liuzhb26\}@mail2.sysu.edu.cn, \IEEEauthorrefmark{3}wuhefeng@gmail.com, \IEEEauthorrefmark{4}linliang@ieee.org}
}

\maketitle

\begin{abstract}
In this paper, we study video synthesis with emphasis on simplifying the generation conditions.  Most existing video synthesis models or datasets are designed to address complex motions of a single object, lacking the ability of comprehensively understanding the spatio-temporal relationships among multiple objects. Besides, current methods are usually conditioned on intricate annotations (\eg~video segmentations) to generate new videos, being fundamentally less practical. These motivate us to generate multi-object videos conditioning exclusively on object layouts from a single frame. To solve above challenges and inspired by recent research on image generation from layouts, we have proposed a novel video generative framework capable of synthesizing global scenes with local objects, via implicit neural representations and layout motion self-inference. Our framework is a non-trivial adaptation from image generation methods, and is new to this field. In addition, our model has been evaluated on two widely-used video recognition benchmarks, demonstrating effectiveness compared to the baseline model. 
\end{abstract}


\section{Introduction}
\label{sec:intro}
Recent developments on video generation have enabled us to synthesize motion-coherent videos, by building variants on the generative adversarial network (GAN)~\cite{goodfellow2020generative}.  The popularities are usually gained on videos of a specific dynamic object (\eg~Tai-Chi-HD dataset \cite{siarohin2019first}), or of a specific action domain (\eg~Kinetics-serious dataset \cite{carreira2018short}).  However, real-life scenes mostly involve multiple distinct instances with multiplex locations with dynamics.  Other than that, there is undoubtedly an increasing demand to make the generation more tractable with a relatively simple generation condition, as an evidence, one of the recent trends of image generation is to condition only on scene graphs \cite{herzig2019canonical}.  To achieve the above two goals, we seek to generate multi-object videos conditioned exclusively on layouts in a single frame. 

A trivial method is to use current layout-to-image models~\cite{li2018layoutgan,zhao2019image}  in a frame-by-frame manner, omitting significant amount of temporal information implicit in the training video, and perhaps requiring supervisions in every frame.  As a rather advanced solution, we may encode the dynamics of object locations together with static object features and other semantics to avoid over supervision, while making the inference of multi-object motions more straightforward and flexible.  See Fig.\ref{fig:first} for a brief illustration.
 \begin{figure}
    \centering\includegraphics[width=0.5\textwidth]{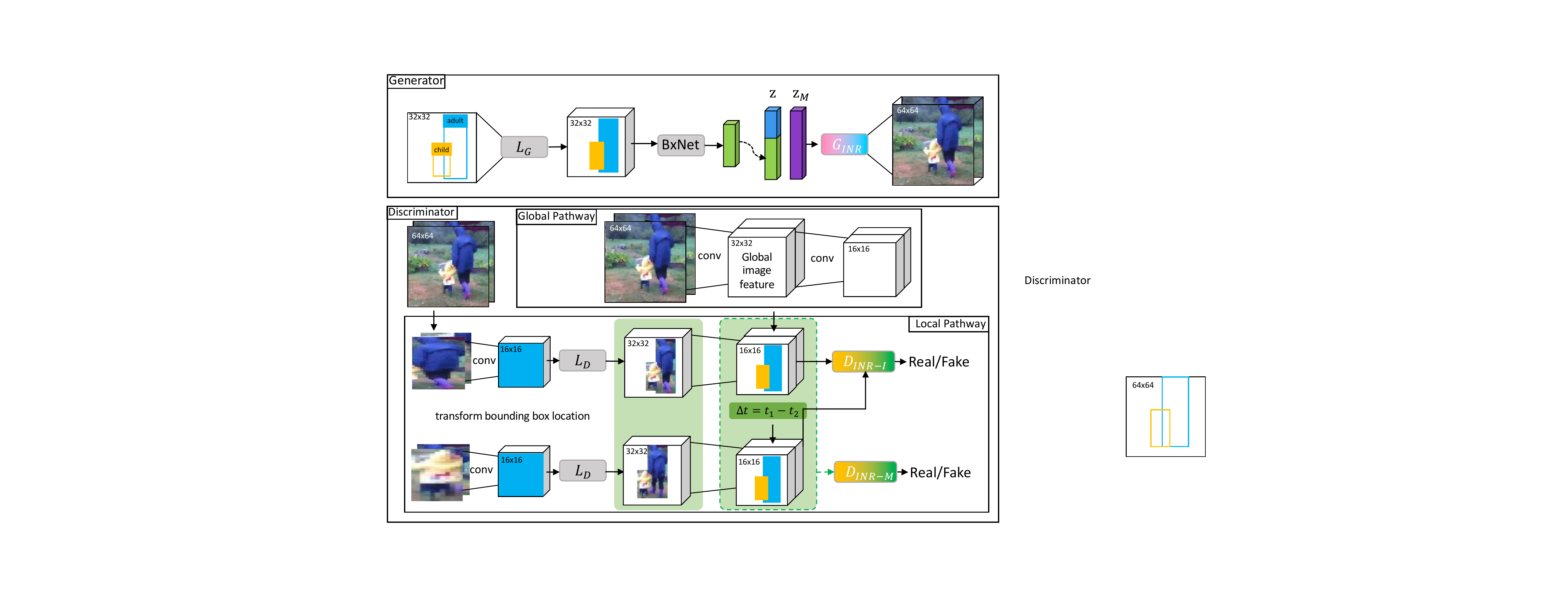}
    \caption{A schematic diagram of multi-object scene video synthesis with our model MOVGAN.}
    \label{fig:first}
\end{figure}
We build up our model --- a \textbf{M}ultiple \textbf{O}bject \textbf{V}ideo \textbf{G}enerative \textbf{A}dversarial \textbf{N}etwork (MOVGAN) --- based on implicit neural representations (INR)~\cite{skorokhodov2021adversarial}, for it offering powerful differentiable continuous signal representations, which in our case, is able to map the generated sequential layout coordinates to high-level features.  To achieve the multiple instances motion inference, similar to layout-to-image tasks~\cite{hinz2018generating}, both the discriminator and the generator of MOVGAN are incorporated with spatial transform neural (STN)~\cite{jaderberg2015spatial} layers to embed a relatively larger amount of object locations.  From model architecture perspective, perhaps the most related works are the recent INR-based GAN~\cite{yu2022generating,skorokhodov2021adversarial} for video generation, since our model can be seen as a generalization from theirs, by embedding instance-level layouts into both the discriminator and generator.

Our contributions can be summarized as follows:
\begin{itemize}
    \item We are the very first kind to study video synthesis based exclusively on layouts from a single frame, and this is useful in some certain cases.
    \item Under our setup, the dynamics of multi-object locations are self-inferred from a single frame, and are dealt separately from other semantics before amalgamation, providing flexibility in modeling object motions.
    \item We have implemented our model on more general video recognition benchmarks (VidVRD~\cite{shang2017video} and VidVOR~\cite{shang2019annotating,thomee2016yfcc100m}) rather than specially designed action video datasets such as VoxCeleb~\cite{Nagrani17}, which is rarely seen in previous works.
\end{itemize}

\section{Related Work}
\label{sec:rltwrk}
%
   
\noindent\textbf{Layout-to-image Generation.}\\
Since a video can be naively treated as an image sequence, image generation methods~\cite{goodfellow2020generative,kingma2019introduction} have an unavoidable enlightenment on video generation tasks.  One of the most noticeable goal of image generation is to attain better controllability even for fine details \cite{sangkloy2017scribbler,singh2019finegan,men2020controllable}.  Such researches usually resort to auxiliary supervisions such as instance-level segmentation masks~\cite{liu2019learning}, bounding boxes~\cite{sun2019image,sylvain2021object}, textual descriptions~\cite{ramesh2021zero} and image layouts~\cite{li2018layoutgan,zhao2019image}.  Among them, coarse layouts (includes bounding boxes and categories) were recognized as perhaps the most flexible and controllable mechanism.  Motivated by this, we seek to generalize from layout-to-image generation to layout-to-video generation.\\

\noindent\textbf{Video Synthesis.}\\
Most video synthesis prior works were task-driven, and can be broadly categorized with respect to the object quantity in the scene --- single or multiple.  Single object tasks favour strongly restricted conditions (\eg~a supplementary source image or video from the same domain) as their semantic source, for example, the talking head generation~\cite{zakharov2019few,chen2020talking}, image animation~\cite{siarohin2019first,sarkar2020neural} and occlusion removal~\cite{pizzati2020model}.  On the other hand, multi-object tasks are more challenging, and commonly apply weakly restricted conditions, such as action labels and language descriptions~\cite{yu2022generating,ge2022long}.  Nonetheless, to better control the synthetic video, recent multi-object tasks intend to bring in strong restrictions such as segmentations~\cite{mallya2020world} and scene graphs~\cite{Cong2022SSGVSSS}, to perform under a video-to-video paradigm .  Our method, instead, detaches multiple layout motions from the overall framework and enjoys freedom for modeling object motions in a simple layout-to-video paradigm.

\begin{figure*}[t]
   \centering
     \begin{subfigure}[b]{0.955\textwidth}
         \centering
          \includegraphics[width=\textwidth]{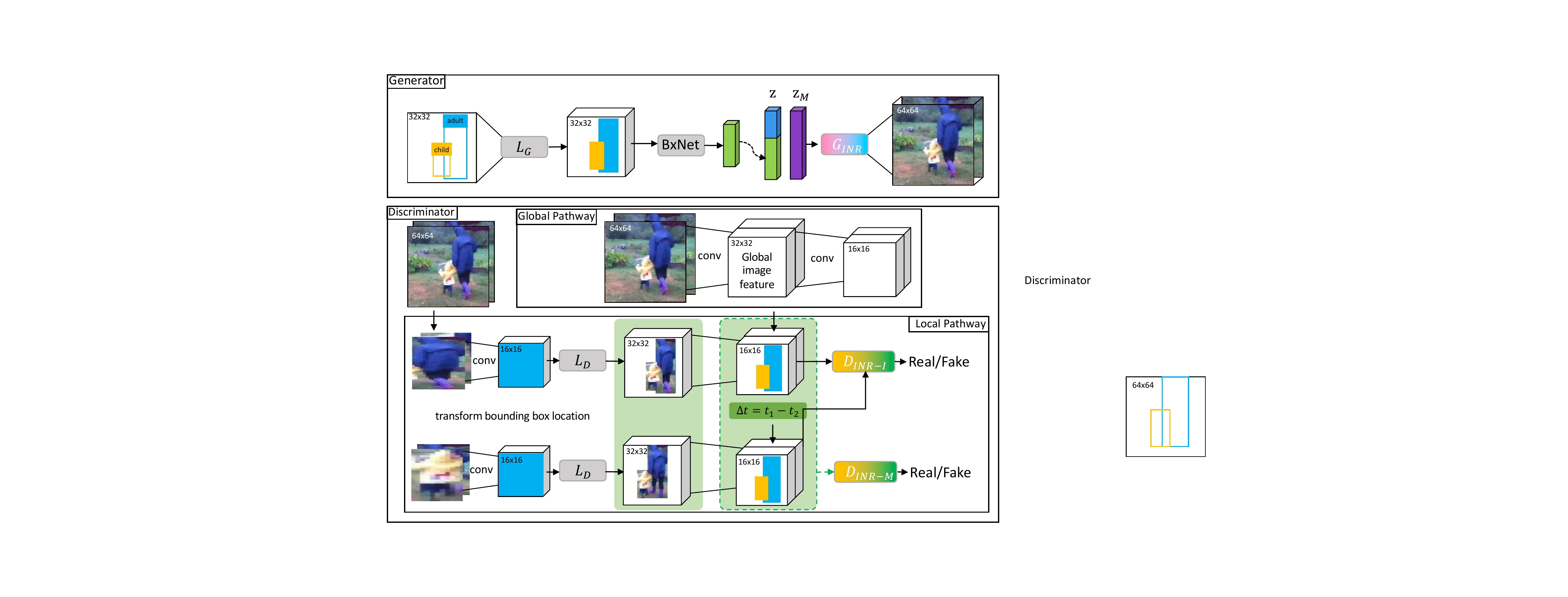}
         \caption{Layout-to-video INR Generator 
         }
         \label{fig:ppl_g}
     \end{subfigure}
   \begin{subfigure}[b]{0.955\textwidth}
         \centering
          \includegraphics[width=\textwidth]{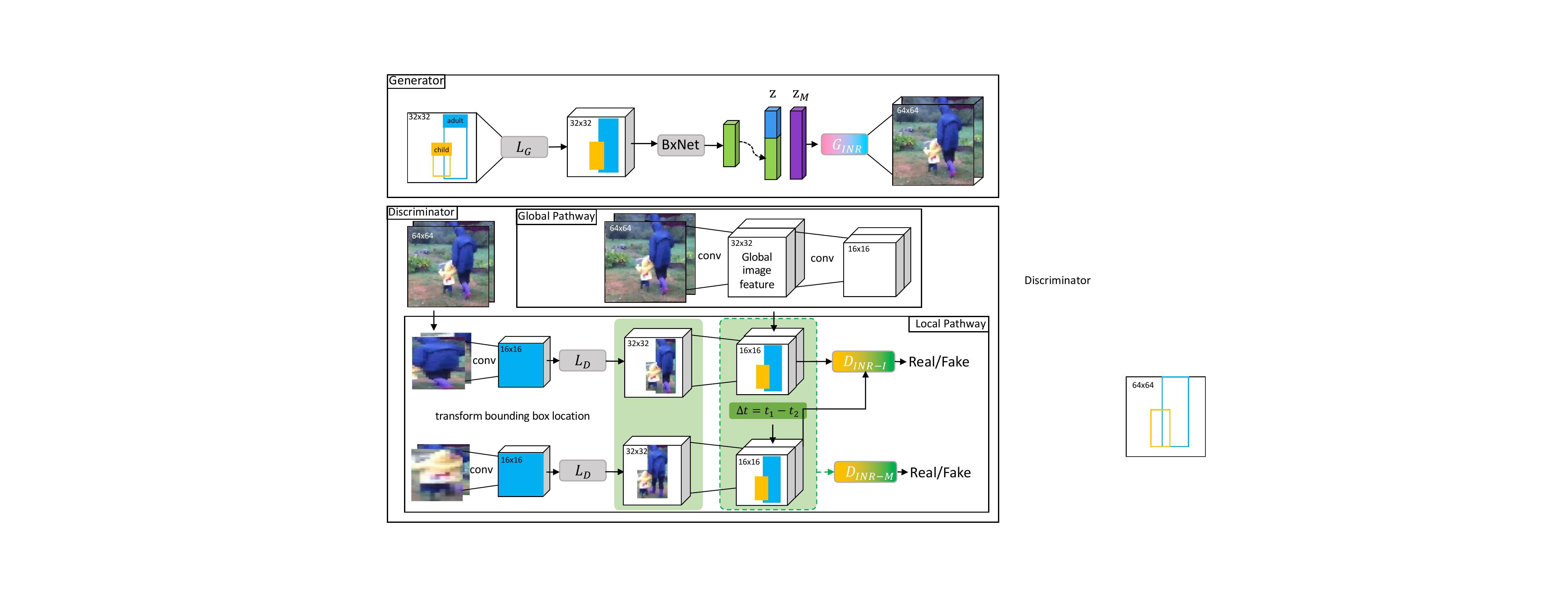}
         \caption{Visual Motion Discriminator}
         \label{fig:ppl_d}
     \end{subfigure}
    \caption{Overall adversarial learning architecture of MOVGAN. The sub-figure (a) illustrates the generator and the bottom (b) shows the inner structure of discriminator. MOVGAN can be treated as an integration of six modules (I)-(VI).}
    \label{fig:pipeline}
\end{figure*}

\section{Methodology}
\label{sec:method}
In this section, we will elaborate our generative framework by first introducing some necessary notations, then describing the major components, and finally presenting the overall learning formula. Our model MOVGAN is built upon DiGAN \cite{yu2022generating} with layout specifics, where both the generator and discriminator are designed to embody layout-encoding modules~\cite{hinz2018generating}. The overall generative backbone is a typical conditional sequential GAN  \cite{wang2020imaginator}, incorporating one latent video generator and a discriminator with two heads to respectively examine image and video authenticity.\\

\noindent\textbf{Model Setup.} \\
Mathematically, our goal is to approximate the real video distribution $P_{\text{real}}$ with a video model distribution $P_G$. 
Following the settings of~\cite{yu2022generating}, we assume a video $\mathbf{v}$ is a sample from $P_G$, \ie~$\mathbf{v}\sim P_G$.  The video $\mathbf{v}(\cdot):\mathbb{R}^3\rightarrow\mathbb{R}^3$ is 3-dimensional mapping from spatial coordinates $(x,y)\in\mathbb{R}^2$ and time $t\in\mathbb{N}^+$ of frame $\mathbf{i}_t$ to the corresponding RGB signals: $\mathbf{v}(x,y,t)=(r,g,b)$.  Conditioning on a multi-layer perception (MLP) with parameter $\phi$, the INR is denoted as $\mathbf{v}(x,y,t;\phi)$.  Given a video of size $H\times W\times T$, we can decode the INR $\mathbf{v}(\cdot;\phi)$ by computing the values of predefined coordinate grids (possibly equally-spaced). In this sense, the INR parameter $\phi$ is generated from the generator $G(\cdot)$ with latent variable $\mathbf{z}$ (typically distributed as Gaussian): $\phi=G(\mathbf{z})$.  The mapping $G(\mathbf{z})$ is treated as a generator in GAN, being adversarially learned together with a discriminator $D(\mathbf{v})$. \\

\noindent\textbf{Layout-to-video INR Generator.} \\
We depict the complete pipeline of our generator in Fig.~\ref{fig:ppl_g}, as can be seen, it contains two streams --- a global and a local pathways --- respectively being designed for global scene and local objects generation.

In the global pathway (I), $G$ takes a random latent vector, the per-frame texture vector $z_{I}$ and the object layout coordinates $b$, as inputs to generate global video canvas feature $f_g$. Global function in (I) processes the labels of each $b_i$ and replicates them spatially to their locations.  In fields where the bounding boxes are overlapped, the label embeddings $l_i$ are wrapped up, whereas the areas with no bounding boxes are filled with zero.  Convolutional layers are then applied to each layout to obtain a high-level layout encoding, thereupon concatenated with $z_I$ to generate the image global feature $f_g$. 

The local pathway (II) controls the generation of local features $f_l$ by merely taking object identities as inputs. Each $f_l^i$ will be further transformed sequentially to locate at the corresponding layout location $b_i$ in the canvas through STN layers. The areas outside layout boxes will still remain zero.  The two streams are then combined together to concatenate the global and local features, $f_g$ and $f_l$, along the channel axis to be synthesized and up-scaled into the final resolution.

The dynamic synthetic module (V) generally follows the INR generation process. Denote the final implicit feature map as $f$, we have $f=\sigma_x\mathbf{w}_x x+\sigma_y\mathbf{w}_y y+\sigma_t\mathbf{w}_t t+\mathbf{b}$, where $\mathbf{w}_x$,$\mathbf{w}_y$,$\mathbf{w}_t$ and $\mathbf{b}$ are the weights and biases of the first layer in the dynamic synthetic module (V); $\sigma_x$, $\sigma_y$, $\sigma_t>0$ are the frequencies of coordinates $(x,y,t)$ in video INR $\mathbf{v}(\cdot;\phi)$.  Note that only the term $\sigma_t\mathbf{w}_t t$ is viewed as a continuous trajectory over time and is determined by parameters excluding $\mathbf{w}_t$, $\phi_t=\phi\backslash\{\mathbf{w}_t\}$. In this sense, we can simply split the combination neural layers into two modules, $G_I$ and $G_M$, so that $G_I$ manages the synthesis of static object/background and $G_M$ animates the outcome of $G_I$. As the input of $G_M$, $z_M$ is the latent vector that squeezes the layout coordinates and/or motions, namely, $G_M(z_M)$ is defined as coordinates motion feature $f_m$. \\

\noindent\textbf{Visual and Motion Discriminator.}\\
Our discriminator is formulated to recognize both the visual-cohesion and the motion-coherent of input videos.  As displayed in Fig.~\ref{fig:ppl_d}, we follow the same spirit in building the generator $G$, that two pathways in the discriminator are configured to extract global/local features.
The obtained global and local features are further concatenated together with the original video and fed into the downstream classifiers.  The two classifiers, denoted by $D_{I}$ and $D_{M}$, respectively identifies static and dynamic visual features.  $D_I$ is a classical image discriminator that takes each frame of the video as input to differentiate its authenticity; $D_M$ is relatively complicated that distinguishes the triplet $(\mathbf{i}_{t_1},\mathbf{i}_{t_2}, \Delta t)$, where $\Delta t:=|t_1-t_2|$ is the time gap between two frames $\mathbf{i}_{t_1}$ and $\mathbf{i}_{t_2}$.  Differing from $D_I$, $D_M$ broadens the input channel from 3 to 7, where the 6th channel is implemented to represent two input video frames and the first channel is for $\Delta t$.  In addition, $D_M$ takes the extracted layout features (both global and local) of frames $\mathbf{i}_{t_1}$ and $\mathbf{i}_{t_2}$ as the input to finely identify the objects' motion by locations. \\

\noindent\textbf{Adversarial Learning.}\\
The learning of our model follows directly from typical conditional GAN under the video generation setting. Considering the coarse layout $L=(b,l)$ with location $b$ and category $l$, the goal of our model is to optimize the following minimax objective function:
\begin{equation}
    \label{eq:cond_gan_obj}
    \begin{aligned}
      \min_G \max_D &V(D,G)=\mathbb{E}_{(\mathbf{v},L)\sim P_{\text{real}}}[\log D(\mathbf{v},L)] \\
      &+\mathbb{E}_{\mathbf{z}\sim P(\mathbf{z}), L\sim P_{\text{real}}}[\log(1-D(G(\mathbf{z},L),L))].  
    \end{aligned}
\end{equation}
In \eqref{eq:cond_gan_obj}, $G(\mathbf{z},L)$ is the integrated generator function and $\mathbf{z} :=(z_I,z_M)$.  Therefore, $G(\mathbf{z},L)$ can be expanded as
\begin{equation}
\label{eq:expand_g}
    \begin{aligned}
      G(\mathbf{z},L)= G_{\text{comb}}(G_I(z_I, L), G_M(z_M)),
    \end{aligned}
\end{equation}
where $G_{\text{comb}}$ stands for the neural function specified in the module (V). $G_M$ here is slightly different from $G_I$ by ignoring the layout $L$, since the object layout/movement information is already embedded in axis motion modeling.  

On the other hand, we split the discriminator function $D$ into $D_I$ and $D_M$ in the visual motion module (VI) (see Fig.~\ref{fig:ppl_d}).  Therefore, $D_I$ and $D_M$ share the layout information conveyed in discriminative pathways (III) and (IV), and we denote $f_t:=D_{\text{layout}}(\mathbf{i}_t, L_t)$ as a function to fuse layout labels $L_t$ with frame $\mathbf{i}_t$ at time $t$.  In this sense, the discriminator function $D(\mathbf{v},L)$ in \eqref{eq:cond_gan_obj} is defined as:
\begin{equation}
\label{eq:expand_d}
    \begin{aligned}
      D(\mathbf{v},L)&=\frac{1}{4}\big[D_I(\mathbf{i}_{t_1}, f_{t_1})+D_I(\mathbf{i}_{t_2}, f_{t_2})\big] \\
      &+ \frac{1}{2}D_M(\mathbf{i}_{t_1}, \mathbf{i}_{t_2},f_{t_1}, f_{t_2}, \Delta t).
    \end{aligned}
\end{equation}
Note that in this equation, we have specified both $L_{t_1}$ and $L_{t_1}$ labels. To avoid confusion, we may explain this with: during the adversarial learning of the discriminator,  $L_{t_1}$ and $L_{t_2}$ are training layout inputs at time $t_1$ and $t_2$, which serve to retain the dynamics between two consecutive frame layouts or object pixels, so that it is consistent with the ground-truth motion pattern.
\section{Experiments}
\label{sec:exp}
The main purpose of this section is to compare MOVGAN with a baseline model, and since there is nearly no algorithm available in our setting, our goal is {\em not} to establish state-of-the-art practical performance.  Instead, these simulation studies serve to verify the effectiveness of our adjustable layout motion inference from four aspects: (1) the capability of generating multi-object videos; (2) the temporal consistency of the generated objects given specific layouts; (3) the visual quality and quantitative scores; (4) the flexibility in editing the generated video.\\
\begin{table}[h!]
\small
\begin{center}

\resizebox{0.475\textwidth}{!}{
\begin{tabular}{|c|c|c|c|c|}
  \hline
  
& \multicolumn{4}{|c|}{{\em VidVRD}}\\
\cline{2-5}
 \multirow{3}{*}[4.5pt]{\textbf{Model}}& \multicolumn{2}{|c|}{FID
 $\downarrow$} & \multicolumn{2}{c|}{FVD$\downarrow$}\\
\cline{2-5}

 &$\times$64&$\times$128&$\times$64&$\times$128\\
 \hline
  TGAN-F&66.23&87.92&665.16&1002.37\\
  Baseline DiGAN & 63.54 & 79.20 & 589.14& 983.74\\
  TGAN-F+ISLA-Norm&60.31&78.03&583.90&881.11\\
  DiGAN+ISLA-Norm&59.02&68.91&579.11&898.62\\
  MOVGAN (ours) & \textbf{57.72} & \textbf{63.52} &\textbf{567.31}&\textbf{877.22}\\
  \hline
   &\multicolumn{4}{c|}{{\em VidVOR} (20 max instance)}\\
\hline
  T-GAN-F& 142.29 & 216.33 & 2445.33 & 3622.71 \\
  TGAN-F+ISLA-Norm & 140.98 & 223.21 & 2536.87 & 3482.28 \\
  Baseline DiGAN & 125.8 & 156.79 &2138.81 & 2309.12 \\
  DiGAN+ISLA-Norm & 114.92&134.82 &2039.2&2156.79 \\
  MOVGAN (ours) & \textbf{101.22} & \textbf{123.78} &\textbf{1781.43} & \textbf{2009.86}\\
  \hline
\end{tabular}
}
\end{center}
\caption{Comparison of FID and FVD on two datasets. 
\label{tab:fid_fvd}} 
\end{table}
\subsection{Datasets} 
We have adopted 2 datasets of progressively increased complexity --- VidVRD \cite{shang2017video} and VidVOR \cite{shang2019annotating,thomee2016yfcc100m}. These two benchmarks were originally created for complex scene video recognition, involving multiple complex scenes, such as indoor and outdoor interactive objects with different light conditions and various camera perspectives.  Most 
importantly, instance-level labels and bounding boxes of each frame are easily accessible.  For all models, we consider a maximum instance number of 11 and 20 respectively for VidVRD and VidVOR; the maximum category number is set to be 36 and 80 respectively for VidVRD and VidVOR.  Further details of these two datasets are listed in Appendix A. \\
\begin{figure}
    \centering
    \includegraphics[width=0.475\textwidth]{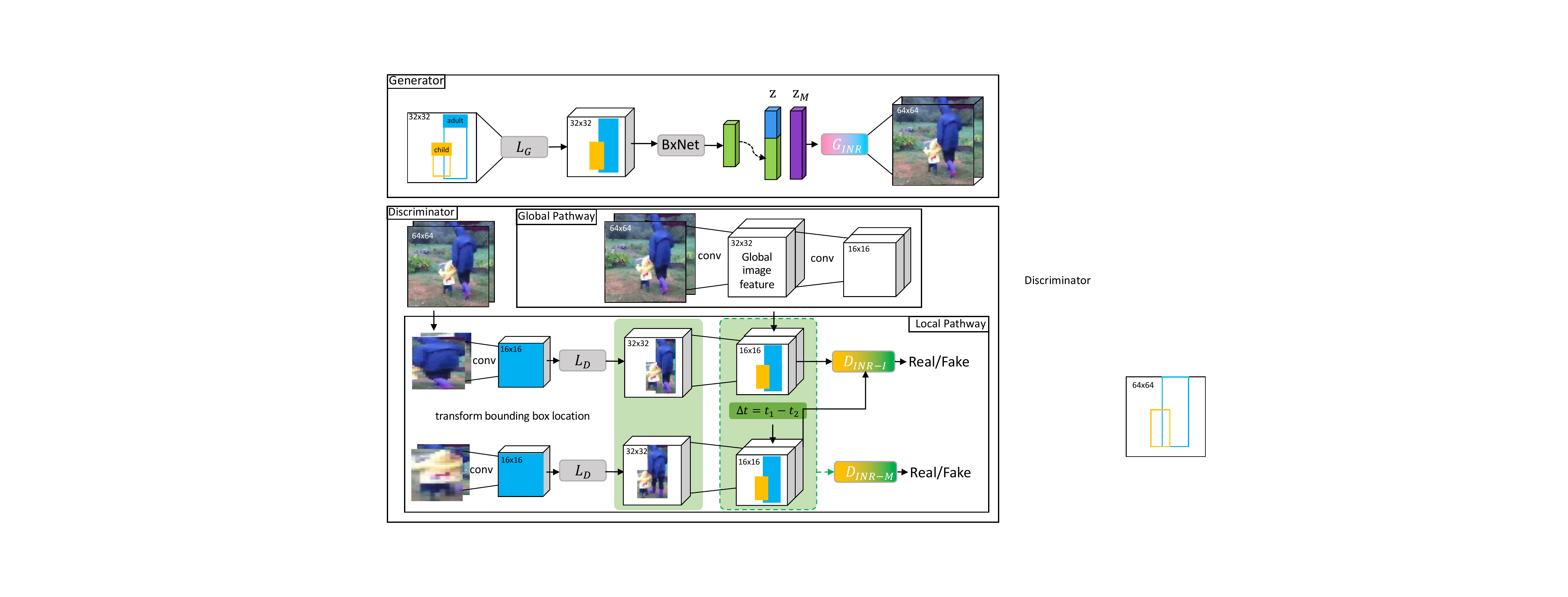}
    \caption{Visualization of editing synthetic results. Each line is two groups of one edited video. Top: adding one cow instance; middle: removing a person instance on the horse; bottom: enlarging both instances.}
    \label{fig:edit}
    \vspace{-10pt}
\end{figure}

 

\begin{figure}[t]
   \centering
     \begin{subfigure}[b]{0.485\linewidth}
         \centering
          \includegraphics[width=0.975\textwidth]{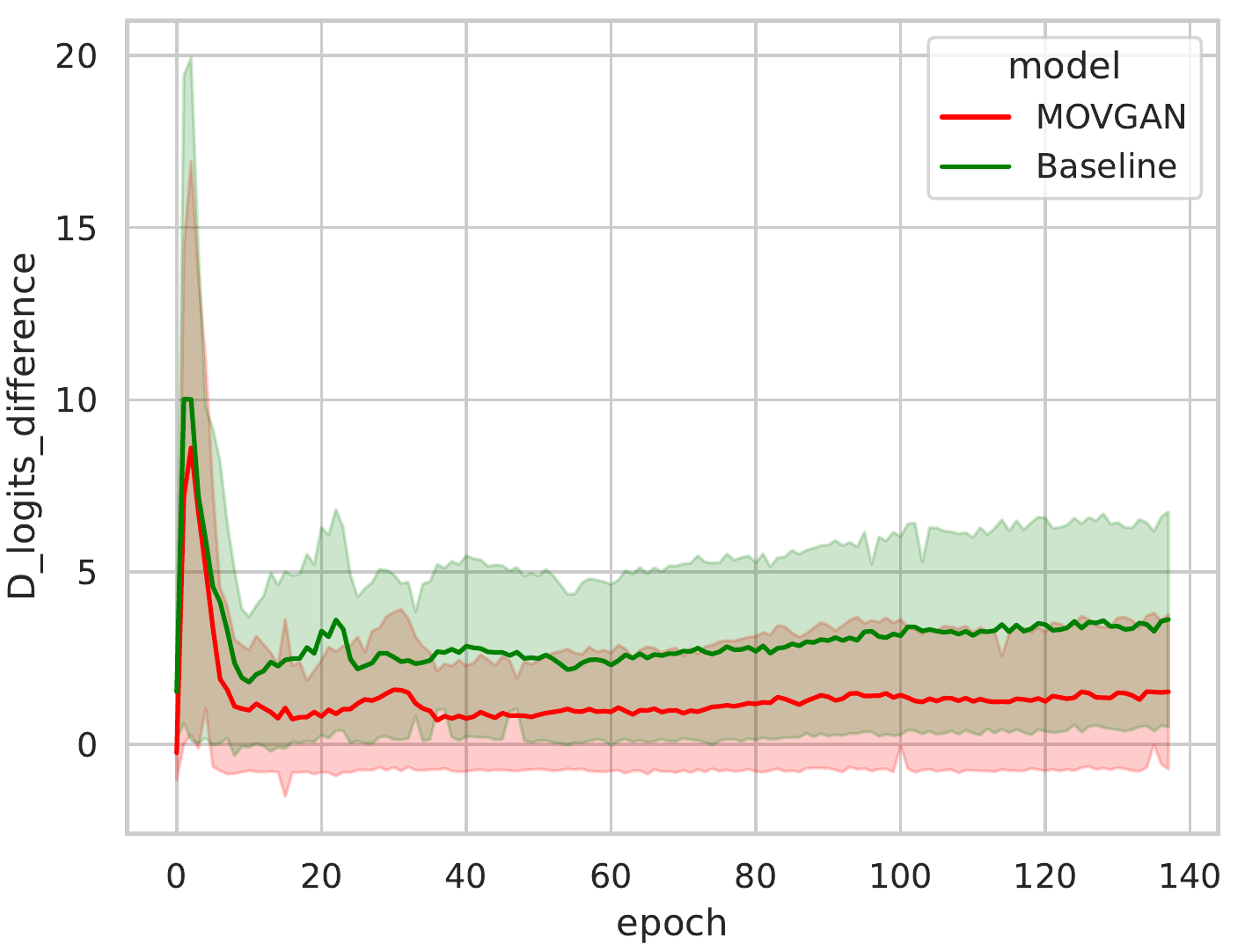}
         \caption{VidVRD}
         \label{fig:logitvrd}
     \end{subfigure}
   \begin{subfigure}[b]{0.485\linewidth}
         \centering
          \includegraphics[width=0.975\textwidth]{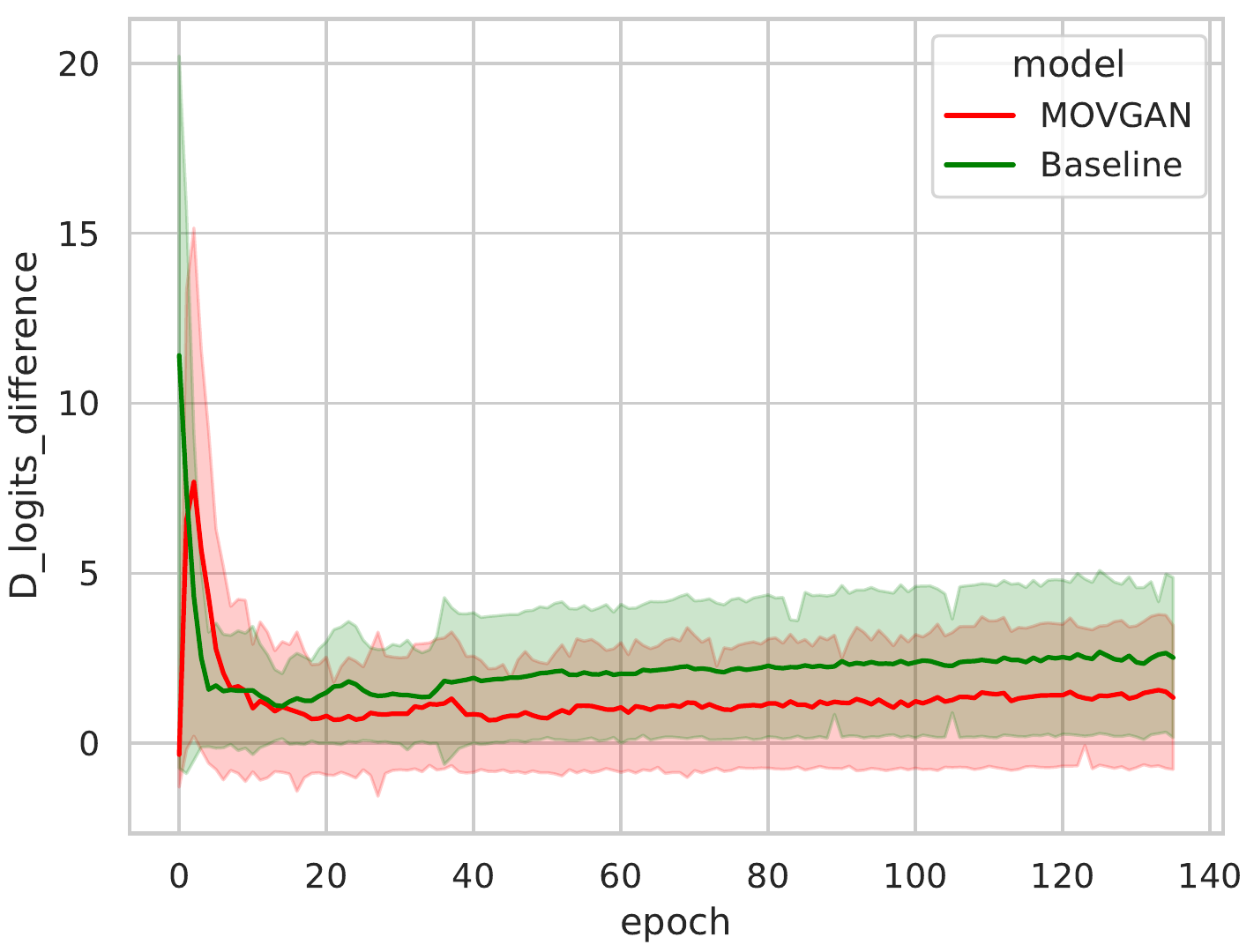}
         \caption{VidVOR}
         \label{fig:logitvor}
     \end{subfigure}
    \caption{Comparison on discriminator logits difference along with the learning iteration.}
    \label{fig:logits}
\end{figure}

\subsection{Baselines and Experimental Settings}
Our baseline model is an adaptation of DiGAN into our setting by conditioning only on multi-object labels (no layouts).  The biggest difference between MOVGAN and the baseline has been shown in Section \ref{sec:method}, in local pathway (II) of generator $G$, where identity embedding layers are introduced.  Except this, the baseline model is maintained nearly identical to MOVGAN for equal comparison.  In practice, those identity embedding layers are implemented based on styleGAN \cite{karras2020analyzing}: the number of linear neural layers of style encoding is set to be 4; the convolutional module is implemented using SkipNet~\cite{wang2018skipnet} without batch normalization.  For more practical reasons, the parameter accuracy of the discriminator is reduced to float16 (default is float32).  Our batch size is 8 with 25,000 frames per epoch,  the learning rate is $5\times10^{-3}$. \\

\subsection{Metrics}
For evaluating per-frame image quality, we measure the Fr\'echet Inception Distance (FID) \cite{heusel2017gans}, whereas for measuring motion-coherency, we choose the Fr\'echet Video Distance (FVD) \cite{unterthiner2018towards}.  For a quick comparison, the inception network of FVD is a specially pretrained inflated 3D Convnet~\cite{carreira2017quo} for sequential data, for the purpose of  estimating distributions over an entire video. \\

\begin{figure*}
    \centering
    \includegraphics[width=0.95\textwidth]{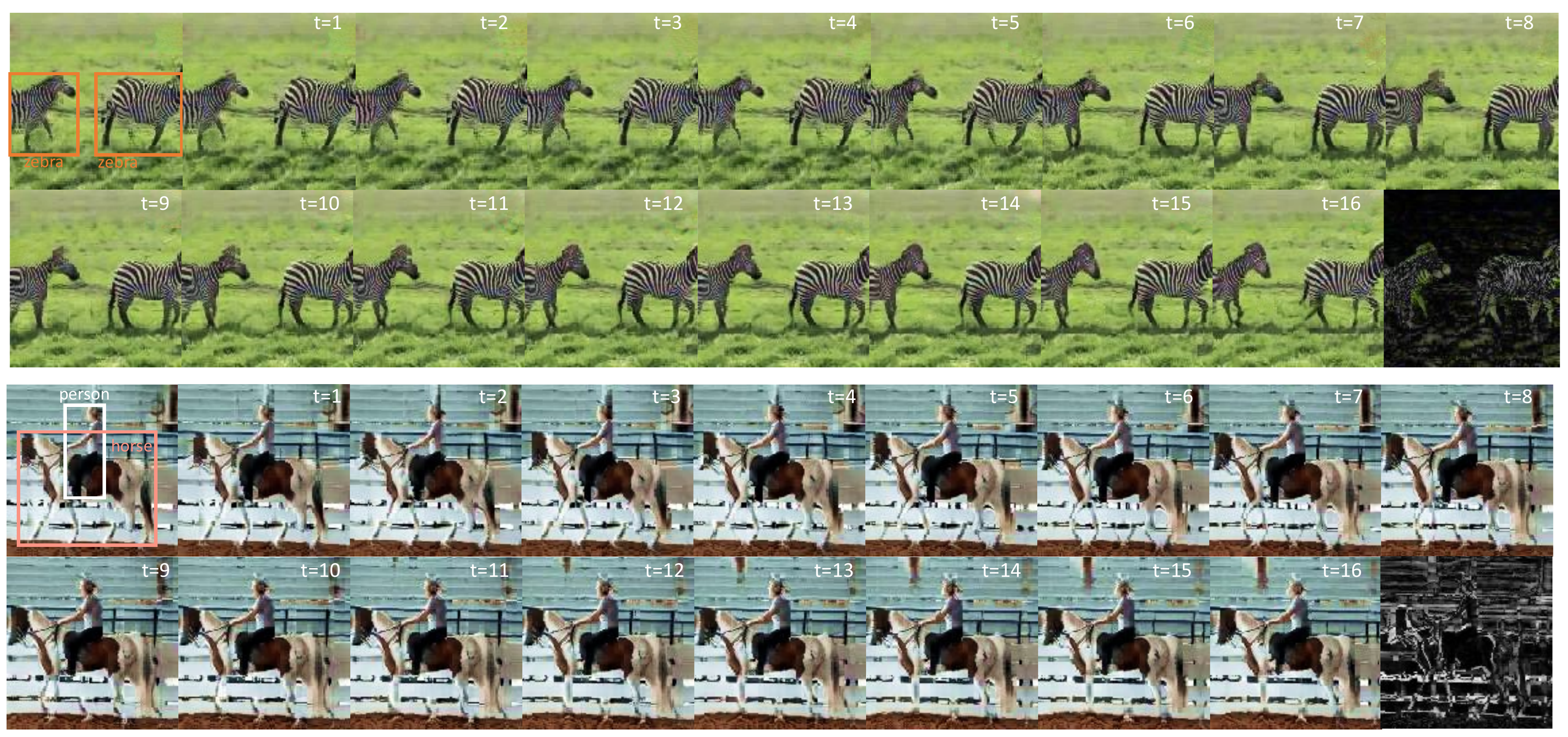}
    \caption{Visualization of generated video clips of 16-frame length. The input layouts are specified at the first frame ($t=1$).}
    \label{fig:vis}
\end{figure*}

\subsection{Quantitative Results}

\begin{figure*}[t]
   \centering
     \begin{subfigure}[b]{0.24\linewidth}
         \centering
          \includegraphics[width=\textwidth]{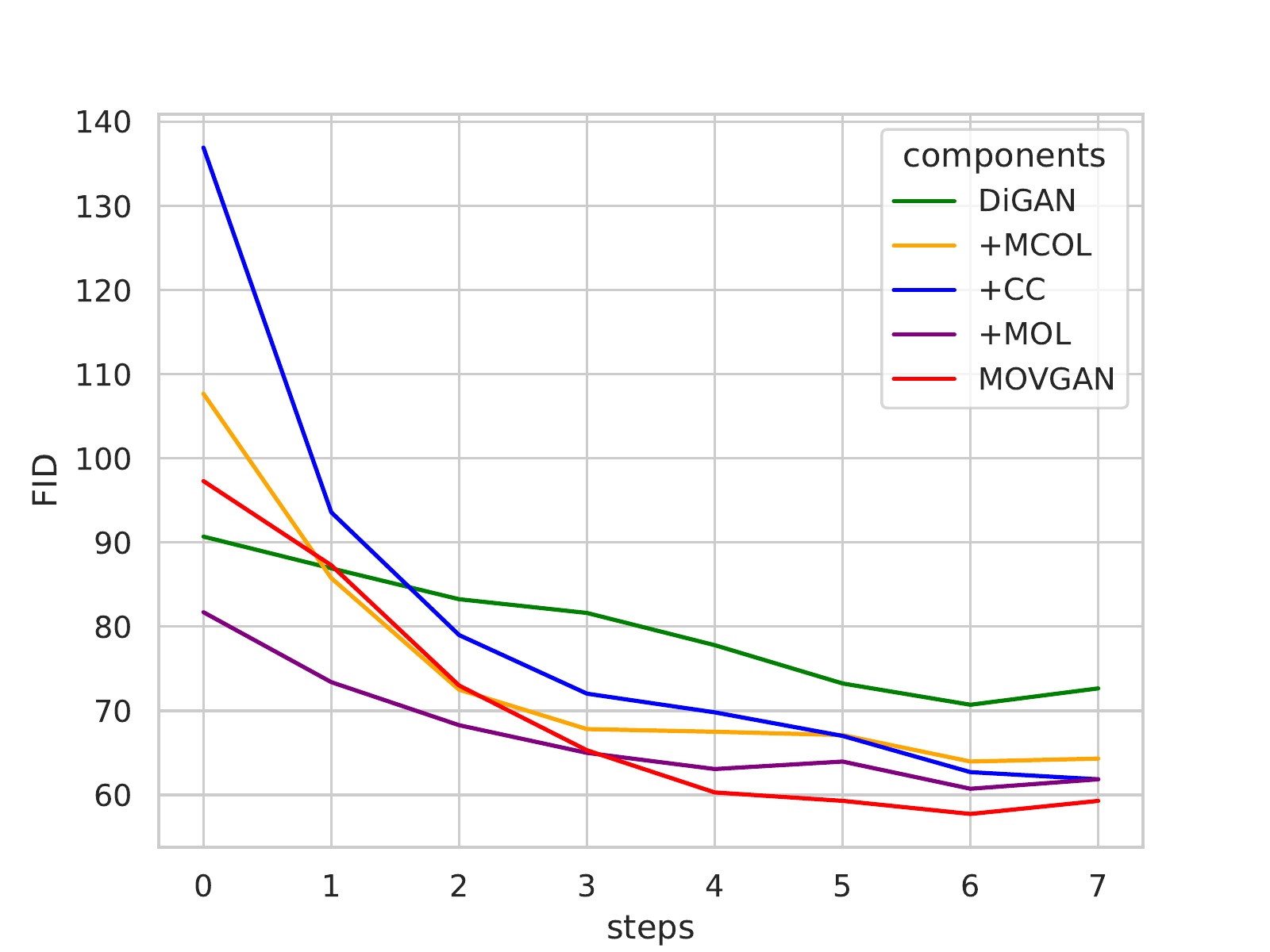}
         \caption{FID records on VidVRD }
         \label{fig:fid_vrd}
     \end{subfigure}
   \begin{subfigure}[b]{0.24\linewidth}
         \centering
          \includegraphics[width=\textwidth]{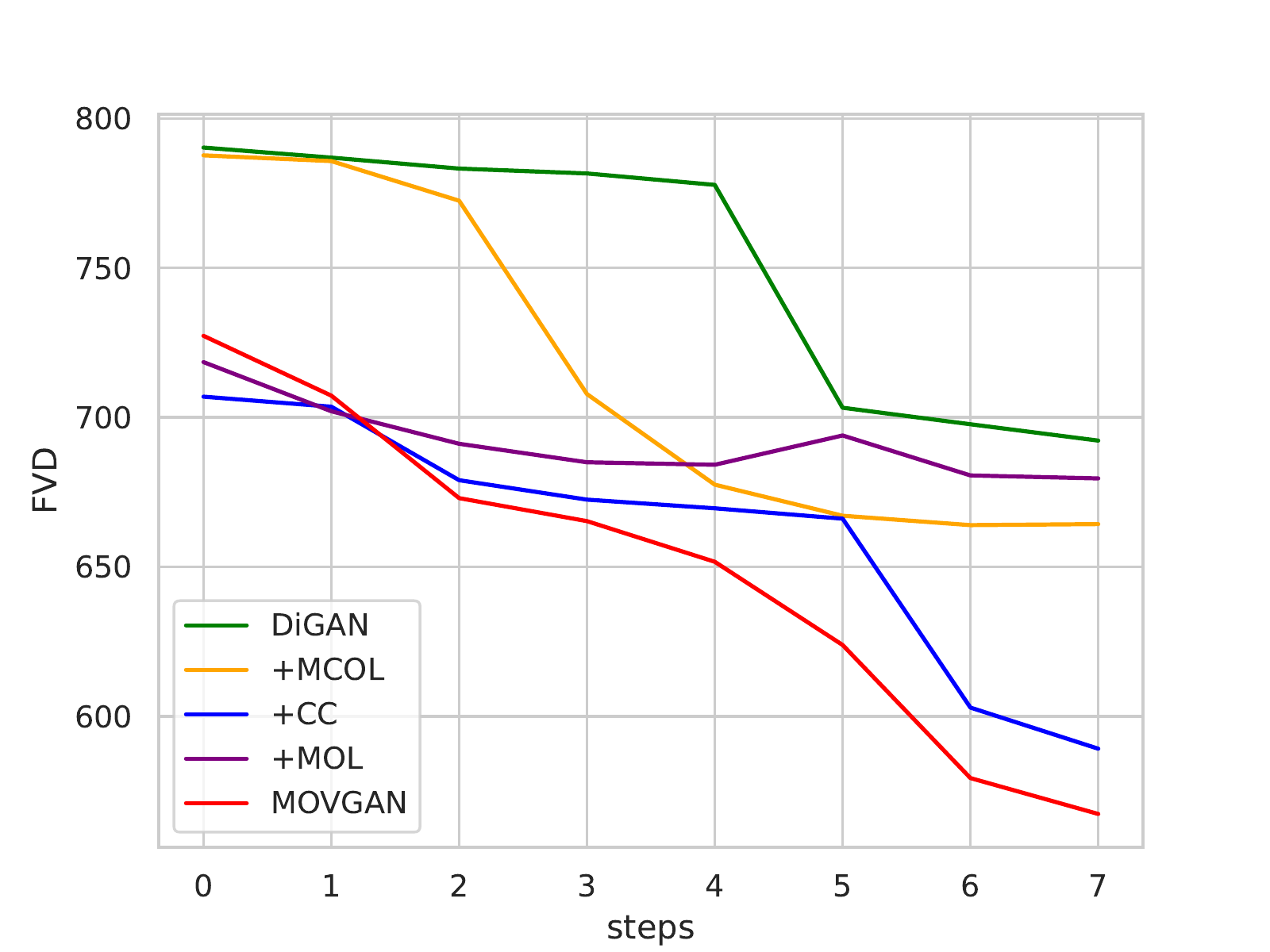}
         \caption{FVD records on VidVRD}
         \label{fig:fvd_vrd}
     \end{subfigure}
     \begin{subfigure}[b]{0.24\linewidth}
         \centering
          \includegraphics[width=\textwidth]{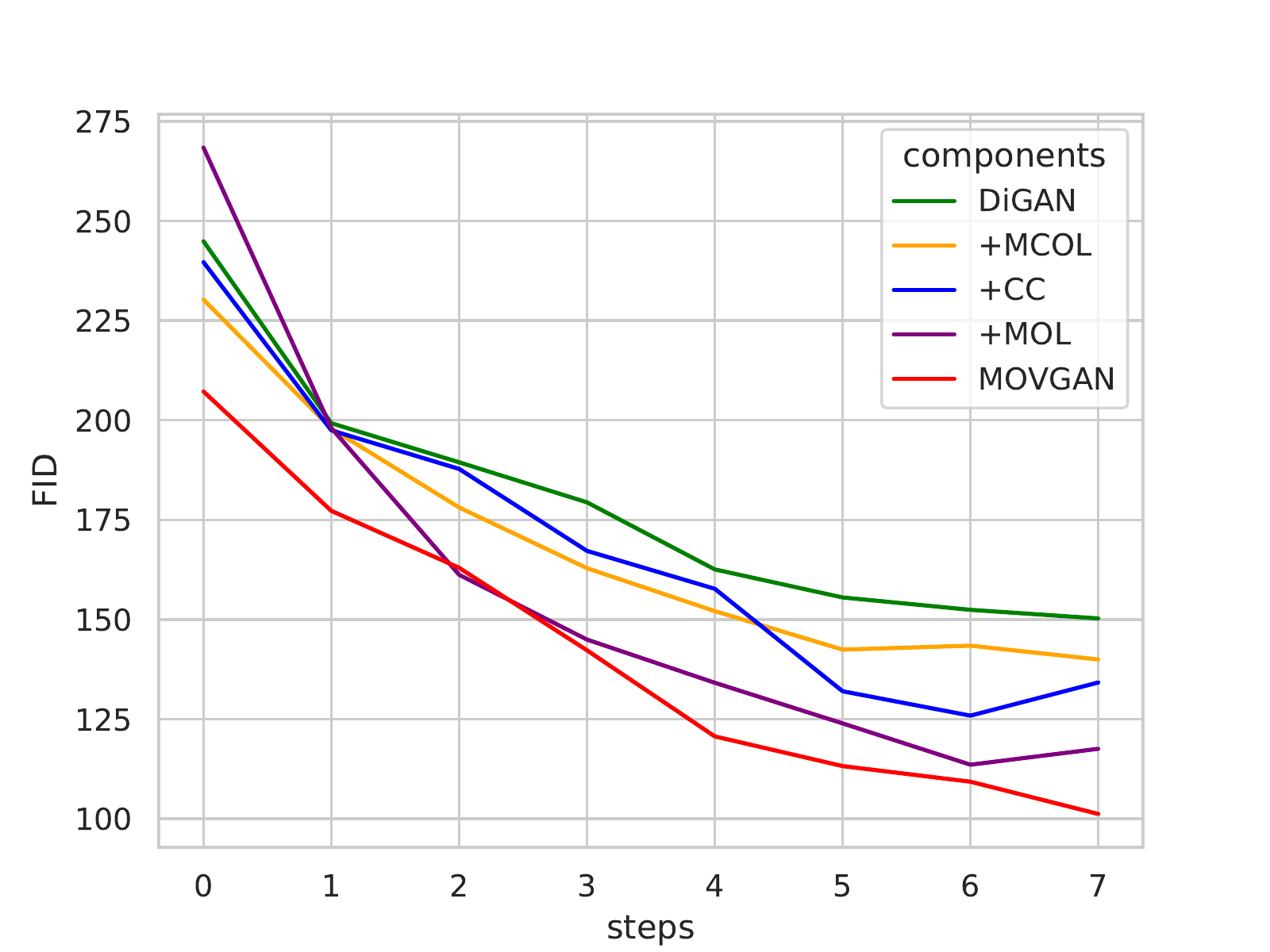}
         \caption{FID records on VidVOR}
         \label{fig:fid_vor}
     \end{subfigure}
   \begin{subfigure}[b]{0.24\linewidth}
         \centering
          \includegraphics[width=\textwidth]{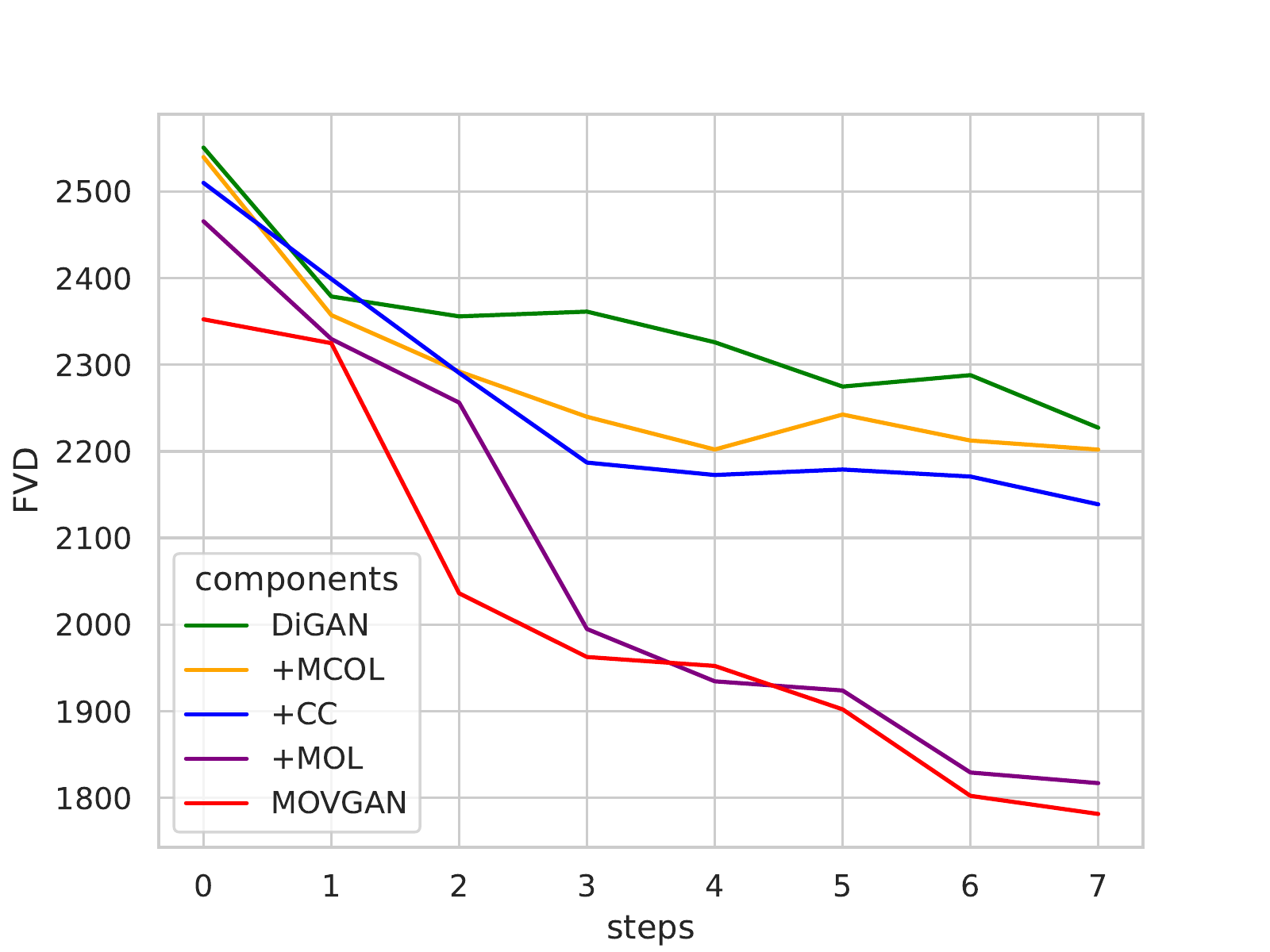}
         \caption{FVD records on VidVOR}
         \label{fig:fvd_vor}
     \end{subfigure}
    \caption{Ablation study results with 64$\times$64 video generation on VidVRD and VidVOR.  The actual step interval in the horizontal axis is recorded every 500 iterations, i.e., 3rd step means 1,500th iteration.}
    \label{fig:ablation}
\end{figure*}
\noindent\textbf{Compared with Baselines}\\
We provide FID and FVD figures for videos of sizes 64$\times$64 and 128$\times$128 in Table \ref{tab:fid_fvd}.  Comparing with the baseline model DiGAN, we see a clear drop of MOVGAN on both scores.  This indicates that layout information embedding indeed helps the model to attain better image quality and motion continuity.  From table \ref{tab:fid_fvd}, TGAN-F \cite{saito2020train} is another video generation baseline. We have also attached a layout embedding approach proposed in \cite{sun2019image}, called ISLA-Norm.  As can be seen that, our layout motion embedding works better than that of the ISLA-Norm in all records.\\

\noindent\textbf{Discriminator Logits}\\
A further evidence of the improvement is displayed through the  discriminator logits curves in Fig.~\ref{fig:logits}.  In particular, the y axis values are calculated between real videos $\mathbf{v}$ and fake videos $\hat{\mathbf{v}}$: for MOVGAN is $\log D(\mathbf{v}, L) - \log D(\hat{\mathbf{v}}, L)$; for Baseline is $\log D(\mathbf{v}, l) - \log D(\hat{\mathbf{v}}, l)$, where one-hot multi-label encoding has been implemented to infuse object identities $l$.  Error bars of each curve indicate ±1 standard deviation computed with respect to the batch mean.  We observe that MOVGAN generally achieves lower discriminator logits, implying better approximations of the real video distribution.\\

\subsection{Alation Study}
Herein, we would like to provide an additional ablative performance evaluation of MOVGAN. The model variants to be measured include:
\begin{itemize}
    \item Action label (DiGAN, green curve)
    \item Multiple-class-object label (+MCOL, yellow curve)
    \item Centralized crop (+CC, blue curve)
    \item Multiple-object layout (+MOL, purple curve)
    \item Multiple-object layout with object identification loss (equation \eqref{eq:cond_gan_obj}, MOVGAN, red curve)
\end{itemize}
For a closer look, we have displayed some of the results in Figure \ref{fig:ablation}, where we have used 5 curves of different colours to indicate the above mentioned variables.  We observe that, by step-wisely applying each of the components to the baseline, FID and FVD gradually converge to lower values.  In particular, we find object layouts (+MOL) are especially beneficial for FVD, which is indeed the case that layout information is crucial in video synthesis.  Henceforth, we may safely conclude that each of the applied component in MOVGAN is effective in their own sense.
\subsection{Intriguing Property}
\noindent\textbf{Visualizations and Video Manipulation.} \\
In Fig.~\ref{fig:edit}, from the top row to the bottom row, we provide video editing examples of three types: `adding' a `cow', `removing' a `person' and `resizing'  two `horses'.  These results have shown the ability of our model in easily operating on fine-level features.  Moreover, in these 6 figures and results shown in Fig.\ref{fig:vis}, from left to right, though being minuscule\footnote{We suggest this is mainly due to the difficulty of the dataset, but since our model is specially designed for multi-objects, it is not our scope to revise our model to fit single object scenarios.  However, it may be of interest to seek refined techniques in future works.}, object motions can be observed. \\

\noindent\textbf{Resolution Augmentation.} \\
We have also managed to furnish an augmentation trick to the current model to rise the resolution of the synthetic videos.  Detailed techniques and results can be found in Appendix B.
\section{Conclusion}
In this paper, we have proposed a new video synthesis framework based on implicit neural representation GAN models, allowing spatio-temporal inference in multiplex scenes.  The notable feature of our model is to depend only on layout information from a single frame and self-inferring the dynamics, opening up the possibility for scenarios with limited supervisions. In future works, it may be of interest to address some of the remaining gaps, such as refining our model or composing complete new methods to achieve greater object movements and more logical interactions.
\section*{Acknowledgment}
This work was supported in part by National Key R\&D Program of China under Grant No. 2021ZD0111601, National Natural Science Foundation of China (NSFC) under Grant No. 61836012 and 62272494, Guangdong Basic and Applied Basic Research Foundation under Grant No. 2023A1515012845 and 2023A1515011374. Also, we want to appreciate Dr. Xu Cai from National University of Singapore for his contributions to this paper. The corresponding author of this paper is Hefeng Wu.
\bibliographystyle{IEEEtran}
\bibliography{IEEEabrv,movgan}

\onecolumn
\appendix

\section{Data Refinement}
\label{spp:dataset}
\textbf{Statistics.} Here we present the detailed statistics of the considered two datasets.  Since these data are now made to serve video synthesis, we seek to optimize the datasets to fit with our experimental setting:
\begin{itemize}
    \item We have taken out possible invalid frames (\eg~those without any object of interest), to avoid data contamination.
    \item We have made the data to be balanced with respect to different object quantities, since majority of the videos only contain two objects.
    \item We have filtered those video clips with their instance amount exceeding the maximum value we set (mentioned in Section~\ref{sec:exp}).
\end{itemize}
Eventually, the statistics of the refined datasets can be found in Table \ref{tab:dataset}.


\begin{table}[h]
\small
\begin{center}
\caption{Dataset statistics.} \label{tab:dataset}
\resizebox{0.475\textwidth}{!}{
\begin{tabular}{|c|c|c|c|c|}
  \hline
 \textbf{Dataset} & \#Video & \#Category 
                & \#Valid Frame & \#Max Instance\\
  \hline
  VidVRD & 1,000 & 35 & 46,930 & 11\\
  VidVOR & 10,000 & 80 & 6,834,925
& 26\\
  \hline
\end{tabular}
}
\end{center}
\end{table}

\begin{table}[h]
\small
\begin{center}
\caption{Scores of FID and FVD after resolution augmentation. 
} \label{tab:fid_fvd_aug}
\resizebox{0.49\textwidth}{!}{
\begin{tabular}{|c|c|c|c|c|}
  \hline
  
& \multicolumn{4}{|c|}{{\em VidVRD}}\\
\cline{2-5}
 \multirow{3}{*}[4.5pt]{\textbf{Model}}& \multicolumn{2}{|c|}{FID
 $\downarrow$} & \multicolumn{2}{c|}{FVD$\downarrow$}\\
\cline{2-5}

 &$\times$128&$\times$256&$\times$128&$\times$256\\
 \hline
  MOVGAN (ours) & 63.52& 72.35 & 877.22 & 983.74 \\
  MOVGAN+ Diff (ours) & 54.67 & 58.26 &890.20& 991.56\\
  \hline
   &\multicolumn{4}{c|}{{\em VidVOR} (20 max instance)}\\
\hline
  MOVGAN (ours) & 123.78 & 153.25 & 1781.43 & 2223.32 \\
  MOVGAN+ Diff (ours) & 112.12 & 124.78 &2002.91 & 2329.36\\
  \hline
\end{tabular}
}
\end{center}
\end{table}


\section{Additional Experimental Results}

\subsection{Resolution Augmentation}
\label{spp:diff_aug}

\begin{figure*}[hb]
    \centering
     \begin{subfigure}[b]{0.475\textwidth}
         \centering
          \includegraphics[width=0.975\textwidth]{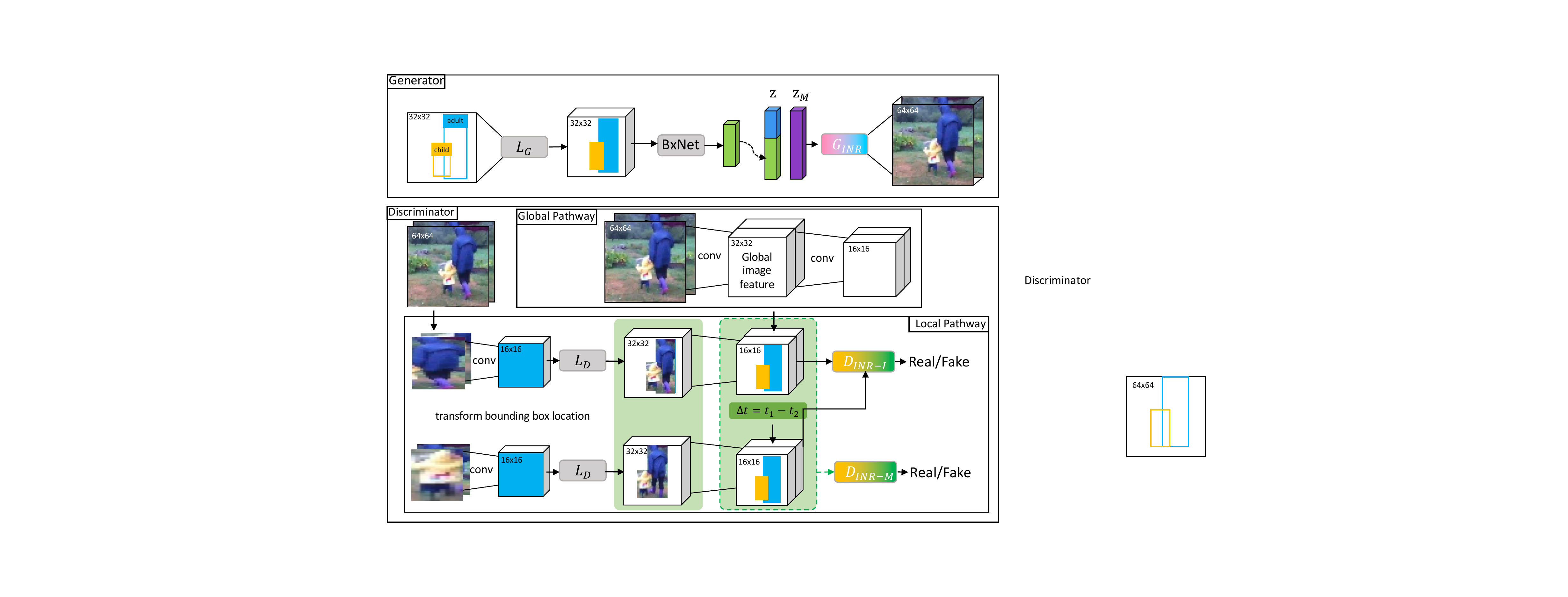}
         \caption{Augmented MOVGAN sample.}
         \label{fig:diff_our}
     \end{subfigure}
     \begin{subfigure}[b]{0.475\textwidth}
         \centering
          \includegraphics[width=0.975\textwidth]{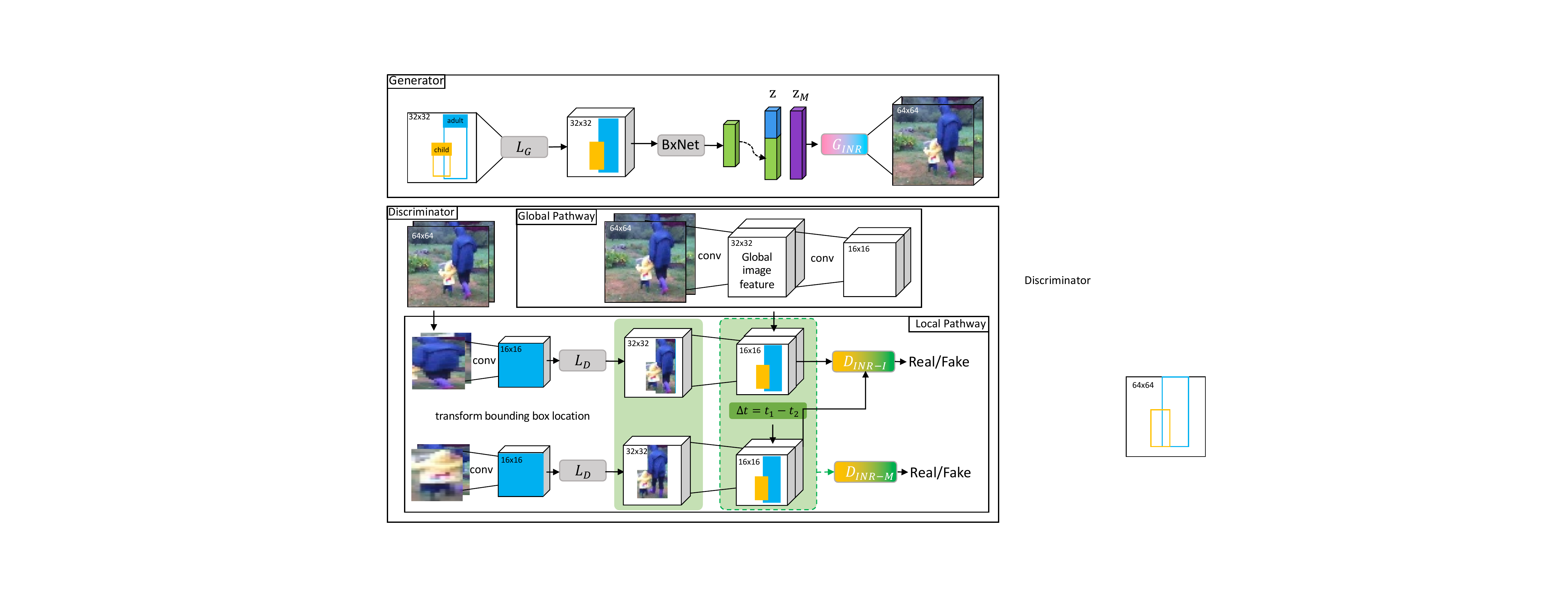}
         \caption{Augmented original training sample.}
         \label{fig:diff_ori}
     \end{subfigure}
    \caption{Comparison of augmented synthetic and training samples.}
    \label{fig:diff}
\end{figure*}
We provide synthetic results in higher resolutions using an augmentation trick, by applying a super resolution algorithm to each synthetic frame.  The algorithm we use is the most recent stable diffusion model~\cite{yang2022diffusion}.  From Fig.\ref{fig:diff}, we see little visual difference between the augmented synthetic and training sample, which is also verified through FID scores in Table \ref{tab:fid_fvd_aug}.  However, at the same time, the FVD score becomes worse, and this might be the reason that some temporal information are missing during augmentation.  Regarding this, it may require more advanced techniques to compensate object dynamics.
\subsection{Further Synthetic Videos}

\begin{figure*}
   \centering
     \begin{subfigure}[b]{0.815\textwidth}
         \centering
          \includegraphics[width=0.975\textwidth]{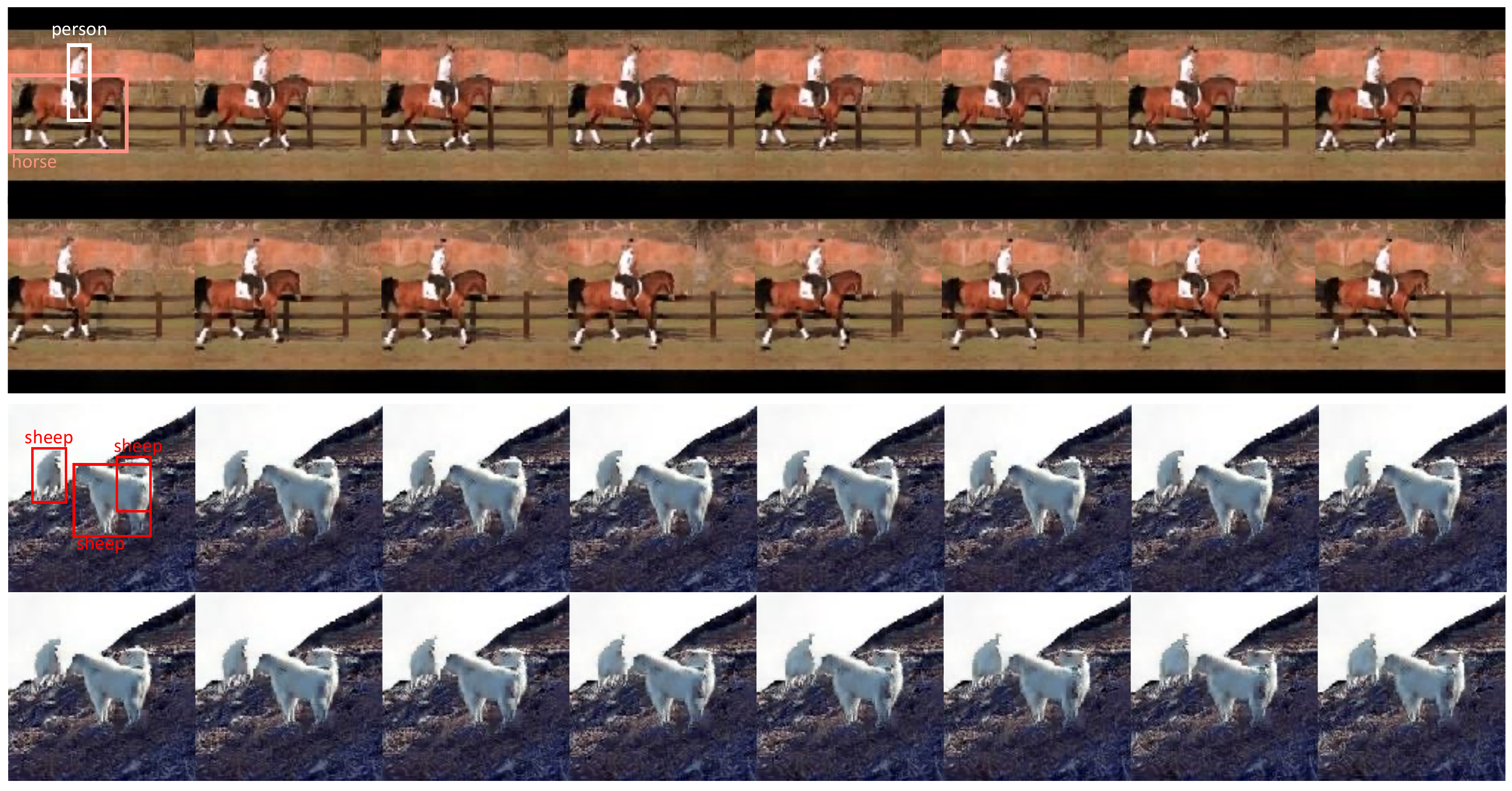}
     \end{subfigure}
   \begin{subfigure}[b]{0.815\textwidth}
         \centering
          \includegraphics[width=0.975\textwidth]{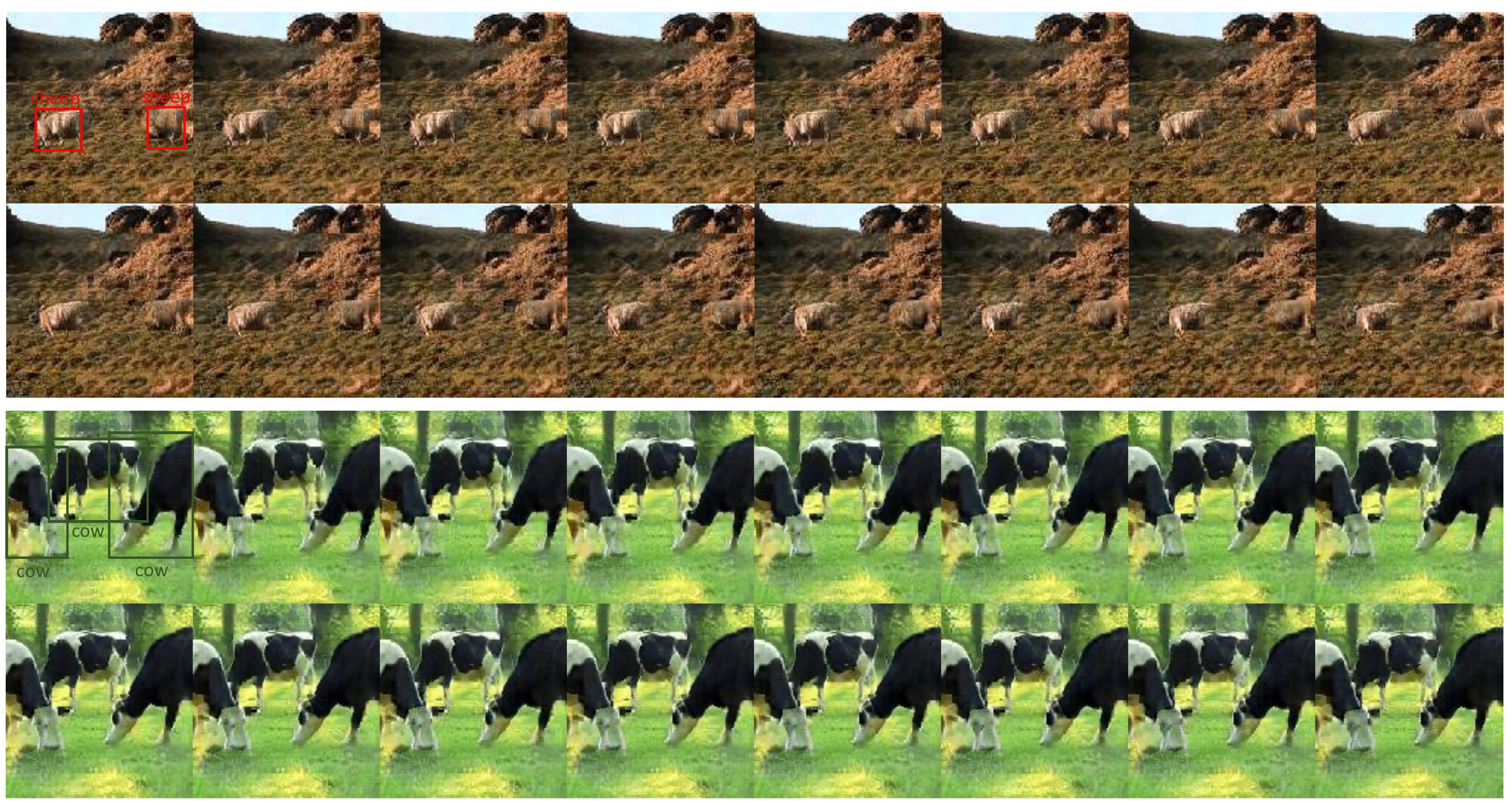}
     \end{subfigure}
    \caption{Further visualizations on synthetic results.}
    \label{fig:supp_vis}
\end{figure*}
\end{document}